\documentclass[preprint,12pt]{elsarticle}

\usepackage[utf8]{inputenc}
\usepackage{soul,color}
\usepackage[margin=1.1in]{geometry}

\usepackage[numbers]{natbib}
\usepackage{mathtools}
\usepackage{amsfonts}
\usepackage{amsmath}
\usepackage{lipsum}
\usepackage{graphicx}
\usepackage{epstopdf}
\usepackage{hyperref}
\usepackage[export]{adjustbox}
\usepackage{float}
\usepackage{tabularx}
\usepackage{multirow}
\usepackage{rotating}
\usepackage[super]{nth}

\DeclareMathOperator*{\argmin}{arg\,min}

\newcommand{\bs}[1]{\boldsymbol{#1}}

\newcommand{\mc}[1]{\mathcal{#1}}

\usepackage[capitalise]{cleveref}
\journal{Elsevier}

\begin{document}
\begin{frontmatter}

\title{An enrichment approach for enhancing the expressivity of neural operators with applications to seismology}

\author[MIT]{Ehsan Haghighat}
\author[KFUPM]{Umair bin Waheed}
\author[BROWN]{George Karniadakis}

\affiliation[MIT]{
    organization={Department of Civil and Environmental Engineering, Massachusetts Institute of Technology},
    city={Cambridge},
    postcode={02139}, 
    state={MA},
    country={USA}
}

\affiliation[KFUPM]{
    organization={Department of Geosciences, King Fahd University of Petroleum and Minerals},
    city={Dhahran},
    postcode={31261},
    country={Saudi Arabia}
}

\affiliation[BROWN]{
    organization={Department of Applied Mathematics, Brown University},
    city={Providence},
    postcode={02912},
    state={RI},
    country={United States}
}

\begin{abstract}
The Eikonal equation plays a central role in seismic wave propagation and hypocenter localization, a crucial aspect of efficient earthquake early warning systems. Despite recent progress, real-time earthquake localization remains challenging due to the need to learn a generalizable Eikonal operator. We introduce a novel deep learning architecture, Enriched-DeepONet (En-DeepONet), addressing the limitations of current operator learning models in dealing with moving-solution operators. Leveraging addition and subtraction operations and a novel `root' network, En-DeepONet is particularly suitable for learning such operators and achieves up to four orders of magnitude improved accuracy without increased training cost. We demonstrate the effectiveness of En-DeepONet in earthquake localization under variable velocity and arrival time conditions. Our results indicate that En-DeepONet paves the way for real-time hypocenter localization for velocity models of practical interest. 
The proposed method represents a significant advancement in operator learning that is applicable to a gamut of scientific problems, including those in seismology, fracture mechanics, and phase-field problems.
\end{abstract}



\begin{keyword}
Earthquake Localization, Neural Operators, DeepONet, machine learning
\end{keyword}

\end{frontmatter}

\section{Introduction}
The Eikonal equation is a cornerstone in a wide range of physical phenomena. It is a fundamental non-linear partial differential equation (PDE) in the domain of geometric optics~\cite{perlick2000ray}, seismology~\cite{nowack1992wavefronts}, computer vision~\cite{rouy1992viscosity}, and Hamiltonian mechanics~\cite{wolansky2013eikonal}. In seismology, the Eikonal equation is used to model wave propagation phenomena, particularly to calculate the travel time of seismic waves through different geological layers. These wavefronts travel at speeds dictated by the material properties of the earth, and the Eikonal equation enables the computation of these varying travel times based on spatially dependent velocities. By accurately characterizing these times, seismologists can infer the properties of subsurface geologies, such as rock type and structure, fluid content, pressure conditions, and stress fields. The Eikonal equation, therefore, plays a crucial role in our understanding of the Earth's interior, aiding in areas from resource exploration to earthquake prediction.

The Eikonal equation is also pivotal in solving the earthquake hypocenter localization problem -- a fundamental task in seismology, encompassing the determination of the initial rupture point or the hypocenter of an earthquake. 
The precise and prompt identification of the earthquake hypocenter holds substantial implications for earthquake early warning systems, rapid response initiatives, and timely seismic hazard assessments~\cite{satriano2011presto}. Advancements in seismic network infrastructure, real-time data processing algorithms, and machine learning have made significant strides in enabling more precise and timely hypocenter localization. However, real-time hypocenter localization remains an elusive goal for seismologists. This requires learning a generalizable Eikonal operator that could be inferred in real-time for predicting hypocenter coordinates corresponding to variations in seismic arrival times and velocity model.

Operator learning, i.e., learning non-linear maps from one infinite-dimensional Banach space to another, seeks to identify the hidden mathematical operators that govern a given dataset or a system, often expressed in terms of PDEs or integral equations. In this context, neural operators have emerged as a promising tool to build surrogate models for obtaining near real-time simulation speeds~\cite{lu2021learning,li2020fourier,li2020neural, lu2022comprehensive,wang2021learning}. 
Neural operators leverage neural networks (NNs) to learn how to map complex relationships within classes of PDEs. They have recently been applied to solve different problems under both data-driven and physics-informed optimizations. Some of the operator learning methods include the Fourier neural operator (FNO)~\cite{li2020fourier}, wavelet neural operator (WNO)~\cite{tripura2022wavelet}, and the graph kernel network (GKN)~\cite{li2020neural}, to name a few. In particular, FNOs have received significant attention and have also been used in seismology for two-dimensional wavefield~\cite{zhang2022learning} and travel time modeling~\cite{suleymanli2023microseismic}. Nevertheless, their applications remain limited to two-dimensional problems as incorporating Fourier transforms in three dimensions is computationally non-trivial.

Built on the foundations of the universal approximation theorem for operators \cite{chen1995universal}, a more general deep neural operator (DeepONet) has been developed by~\cite{lu2021learning} that allows one to fully learn the underlying operator and thus perform real-time predictions for arbitrary new inputs and complex domains. DeepONets consist of two constituent parts: a `branch net' and a `trunk net.' The branch net encodes the input function using a predetermined number of sensors. The trunk net, on the other hand, encodes the continuous solution operator as a function of the spatiotemporal inputs. These two networks collaborate to transform the input data into the designated outputs. The unique architecture of DeepONets enables them to capture and learn complex operators directly from data. Building upon this, Physics-Informed DeepONets (PI-DeepONets) augment the capabilities of DeepONets by incorporating known physical laws or governing equations as additional constraints during the training process~\cite{wang2021learning}. 

Despite its success in learning operators for a variety of computational science and engineering problems~\cite{goswami2022physics,wang2022improved,goswami2022physics2,lanthaler2022error,lutjens2022multiscale,kovachki2023neural,lin2023learning,he2023novel,li2023phase,koric2023deep,garg2023vb,koric2023data}, DeepONets show weakness in learning operators in which the output solution moves spatially, corresponding to variations in the input function (see Supplementary Materials). Therefore, in this paper, we propose a novel DeepONet architecture, namely the Enriched DeepONet (En-DeepONet), that overcomes the aforementioned limitation. This is particularly useful for learning the Eikonal operator, in which we map the observed travel times at seismic stations to travel time values for the entire computational domain. The hypocenter location is then obtained by the coordinates of the minimum travel time value. 
Compared to the standard DeepONet architecture, represented in Fig.~\ref{fig:fig1}-a,
the En-DeepONet (Fig.~\ref{fig:fig1}-b) leverages addition and subtraction operators as well as a newly introduced `root' network for constructing the surrogate model. 
Adding such operations resembles the \emph{enrichment} strategies in classical numerical methods like the Finite Element Methods~\cite{moes1999finite,karniadakis2005spectral}, hence the name of the proposed architecture. 
A comparison of the En-DeepONet architecture against the standard DeepONet architecture is presented in the Supplementary Materials, highlighting up to four orders of magnitude improved accuracy without an increase in training cost or the number of trainable parameters under both supervised and unsupervised (physics-informed) optimizations. 

We apply the proposed framework to earthquake localization under both variable velocity and arrival time conditions that characterize real-world scenarios. We also study the sensitivity of the framework in hypocenter localization with respect to noisy arrival times. Through rigorous testing and evaluations, we observe the remarkable performance of En-DeepONet in achieving near real-time localization capabilities even for highly heterogeneous velocity models. The ability to accurately localize earthquake hypocenters amidst such variability is a significant advancement in the realm of seismology. 

By harnessing the capabilities of the novel En-DeepONet architecture, our work addresses existing limitations and paves the way for more advanced and adaptable solutions in seismology and beyond. With an improved understanding of seismic events and the capability to predict them in near real-time, we can substantially improve the efficiency of early warning systems. This, in turn, contributes to the broader objective of disaster risk reduction, enabling quicker and more informed decisions during an earthquake event to mitigate its impact and save lives.

\section{Problem statement}

\subsection{Eikonal equation}

We cast the source localization task as a traveltime modeling problem. Given first-arrival traveltimes observed at seismic sensors, we build the traveltime field in the entire computational domain using En-DeepONet. Once that is obtained, the hypocenter location is given by the coordinates of the minimum traveltime value.

Given a source location $\bs{x}_s$ and a velocity field $v(\bs{x})$, the Eikonal equation provides the first arrival traveltime field $T(\bs{x})$. The Eikonal equation is expressed as
\begin{equation}
\| \nabla T(\bs{x}) \|^2  = \frac{1}{v(\bs{x})^2}, \quad \text{s.t.} \quad T(\bs{x}_s) = 0, 
\end{equation}
where $\nabla$ is the gradient vector. This PDE forms a nonlinear and hyperbolic first-order partial differential equation that requires special solution methods. This equation implies that the gradient of the arrival time surface is inversely proportional to the speed of the wavefront.

\subsection{Earthquake hypocenter localization using Eikonal equation}

The hypocenter localization problem is solved using the Eikonal equation through an iterative inversion process that seeks to minimize the misfit between observed and calculated traveltimes~\cite{lee2003hypo71}. This process begins with an initial hypocenter location guess and a subsurface velocity model, followed by forward modeling to obtain theoretical traveltimes at seismic stations. The misfit between the calculated and observed traveltimes is then computed, and an optimization algorithm is employed to update the hypocenter location accordingly. By repeating these steps until convergence or a predefined stopping criterion is met, the hypocenter location is refined, ultimately providing a solution closely aligned with observed traveltimes at seismic stations. 

A number of methods have been developed over the years to solve the hypocenter localization problem. Heuristic approaches, like the genetic algorithm~\cite{sambridge1993earthquake} or simulated annealing~\cite{billings1994simulated}, are among the methods used to seek out the hypocenter with the least error. Other approaches involve systematic searching or sampling techniques to gauge the probability of the earthquake location over the model space~\cite{lomax2000probabilistic,lomax2009earthquake}. For instance, grid search methodically covers the model space to calculate the probability distribution of the hypocenter location. However, this approach can be computationally demanding due to the repeated forward solutions and optimization process. Moreover, it is unable to meet the need for real-time hypocenter localization. More recently, several deep learning-based localization methods have been developed that have shown remarkable performance in real-time localization~\cite{wang2021data,wamriew2021deep,izzatullah2022laplace}. However, their velocity generalization capabilities are often in question.


\subsection{Surrogate modeling using DeepONets}

An alternative strategy involves solving the Eikonal equation for potential source locations a priori and constructing a surrogate model that encompasses all possible solutions, using arrival times at seismic stations as inputs. 
If this surrogate model can be accessed and queried in real-time, it enables rapid localization of an earthquake's hypocenter. While conventional neural networks may be employed for this purpose~\cite{anikiev2022traveltime} with limited efficacy, DeepONets have demonstrated exceptional capabilities as a robust and efficient tool for building such surrogate models.

\subsection{DeepONet: A Data-Driven and Physics-Informed Efficient Architecture for Surrogate Modeling}

\begin{figure}
    \centering
    \includegraphics[width=1.\textwidth]{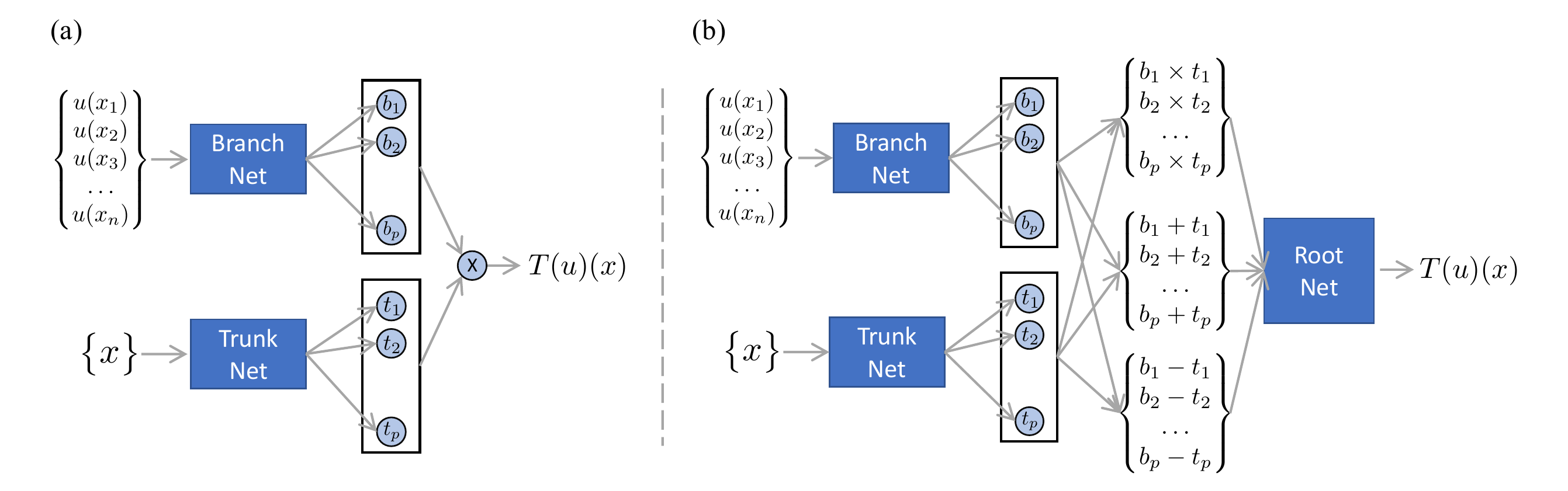}
    \caption{{Illustration of alternative deep neural operators.} (a) The standard DeepONet architecture incorporates two sub-networks, namely the `branch network' and the `trunk network.' The function of the branch network is to encode the input data, leveraging a set number of prearranged sensors. The trunk network is used to encode the locations of the output function. Together, these networks allow learning the transformation of input data into the designated output parameters. (b) The proposed En-DeepONet architecture leverages the addition and subtraction operators, in addition to multiplication, to combine the outputs of branch and trunk networks. We pass this output to a root network that helps to learn more generalized mathematical operators. }
    \label{fig:fig1}
\end{figure}

Operator networks have recently emerged as promising tools for developing surrogate solvers, which can provide instantaneous solutions based on various sensor readings. These sensors refer to measurements of specific field variables at fixed locations. The network architecture consists of two main components: a branch-net, which encodes the sensor data, and a trunk-net, which approximates the continuous solution. The outputs of these networks, known as the embedding nodes, are combined through a dot product to construct the final output or operator. The DeepONet architecture is expressed as~\cite{lu2021learning}: 
\begin{equation}
    T(u)(x, t) \approx \sum_{k=1}^p b_k(u_1, u_2, \dots, u_p) t_k(x, t),
    \label{eq:eq2}
\end{equation}
with $u_i$ as the sensor readings at fixed locations, and $b_k$ and $t_k$ as outputs of branch and trunk networks, as depicted in \cref{fig:fig1}-a. $b_k$ and $t_k$ are embedding bases for the sensor and solution networks. 

Such a network construction can be trained under data-driven (supervised) or data-free physics-informed (unsupervised) approaches. Of course, the training cost of the latter is much more than the former because of Enriched computational graphs due to AD and the more complex loss envelope and multi-objective loss function. Nevertheless, given a set of inputs $Y$ and sensor readings  $U_i$ at fixed locations and target data $G$, we can set up the optimizer using the MSE loss function as follows:
\begin{equation}\label{eqs:deeponet}
    \argmin_{\bs{\theta} \in \mathbb{R}^N} \mc{L} = MSE(T(u)(x, t) - T^*),
\end{equation}
where $\bs{\theta}$ represents the set of all trainable parameters, and $T^*$ represent the expected values for the travel time.
Given the continuous construction with respect to the spatio-temporal input set $y$, we can also leverage AD to incorporate any PDE-based losses, just like the standard PINN (for more details, see \cite{wang2021learning}).

\subsection{En-DeepONet}

Here, we propose a revised DeepONet architecture, namely, the Enriched-DeepONet (or En-DeepONet for short), for traveltime modeling. Standard DeepONet architecture, expressed in \cref{eqs:deeponet}, demonstrated its effectiveness in addressing problems with variable amplitude of the expected solution. However, in the context of earthquake source localization, the solution itself shifts spatially as the hypocenter moves. We find that standard DeepONet performs poorly in capturing such solutions, which is rooted in the use of the projection (multiplication) operator. We also find that adding summation or subtraction operation—i.e., $t_k(x, t) \pm b_k(u_1, u_2, \dots, u_p)$—improves the learning capacity of DeepONets significantly for such problems. Given that this operation does not reduce the output dimension, we add an additional linear layer to reduce the output dimension to a single output, similar to what a dot-product does. This is expressed mathematically as:
\begin{equation}\label{eqs:xdeeponet}
\begin{split}
    T(u)(x, t) &\approx \sum_{k=1}^p w^+_k (t_k(x, t) + b_k(u_1, u_2, \dots, u_p)) \\
               &+ \sum_{k=1}^p w^-_k (t_k(x, t) - b_k(u_1, u_2, \dots, u_p)) \\
               &+ \sum_{k=1}^p w^*_k (t_k(x, t) \cdot b_k(u_1, u_2, \dots, u_p)),
\end{split}
\end{equation}
where, $w_k^\circ$ are summation weights (linear layer) associated with each operation. 
Note that setting $w_k^\pm = 0$ and $w_k^* = 1$ recovers the original DeepONet construction, expressed in Eq.~\eqref{eq:eq2}. 
Finally, we can even further generalize the DeepONet architecture, by adding a full neural network instead of the linear layer, namely RootNet, as shown in \cref{fig:fig1}-b. Therefore, we call this generalized construction the Enriched-DeepONet or En-DeepONet. Therefore, the most general form of En-DeepONet is expressed mathematically as
\begin{equation}\label{eqs:xdeeponet-full}
\begin{split}
    T(u)(x, t) &\approx r\left(
        \begin{Bmatrix}
            t(x, t) \cdot b(u_1, u_2, \dots, u_p) \\
            t(x, t) + b(u_1, u_2, \dots, u_p) \\
            t(x, t) - b(u_1, u_2, \dots, u_p)
        \end{Bmatrix}
        \right),
\end{split}
\end{equation}
where $r(\circ)$ stands for the output "root" network.

\section{Applications}

In this section,  we explore the application of the En-DeepONet architecture for instant traveltime field estimation. To this end, we first explore its application to hypocenter localization of the Marmousi velocity model~\cite{versteeg1994marmousi}, a highly heterogenous velocity model, leading to highly accurate realtime predictions. Furthermore, by considering a range of simple and complex velocity models, we illustrate the capacity of En-DeepONet to generalize over velocity model variations, ultimately harnessing the full potential of the operator learning paradigm. 

We first present a comparison of the En-DeepONet architecture against the  standard DeepONet architecture, highlighting the need for the addition and subtraction operators, in addition to the multiplication operator used in the standard DeepONet construction, to obtain optimal accuracy when the expected solution shifts with the change in sensor position. We then apply the framework to a set of problems with increasing complexity. All problems are set up in SciANN \cite{haghighat2021sciann}, a TensorFlow/Keras API optimized for physics informed machine learning, and codes reproducing the results will be made public at \href{https://github.com/ehsanhaghighat/En-DeepONet}{https://github.com/ehsanhaghighat/En-DeepONet}.

\subsection{Why En-DeepONet?}\label{append:validation}

\begin{figure} 
    \centering
    \includegraphics[width=0.95\textwidth]{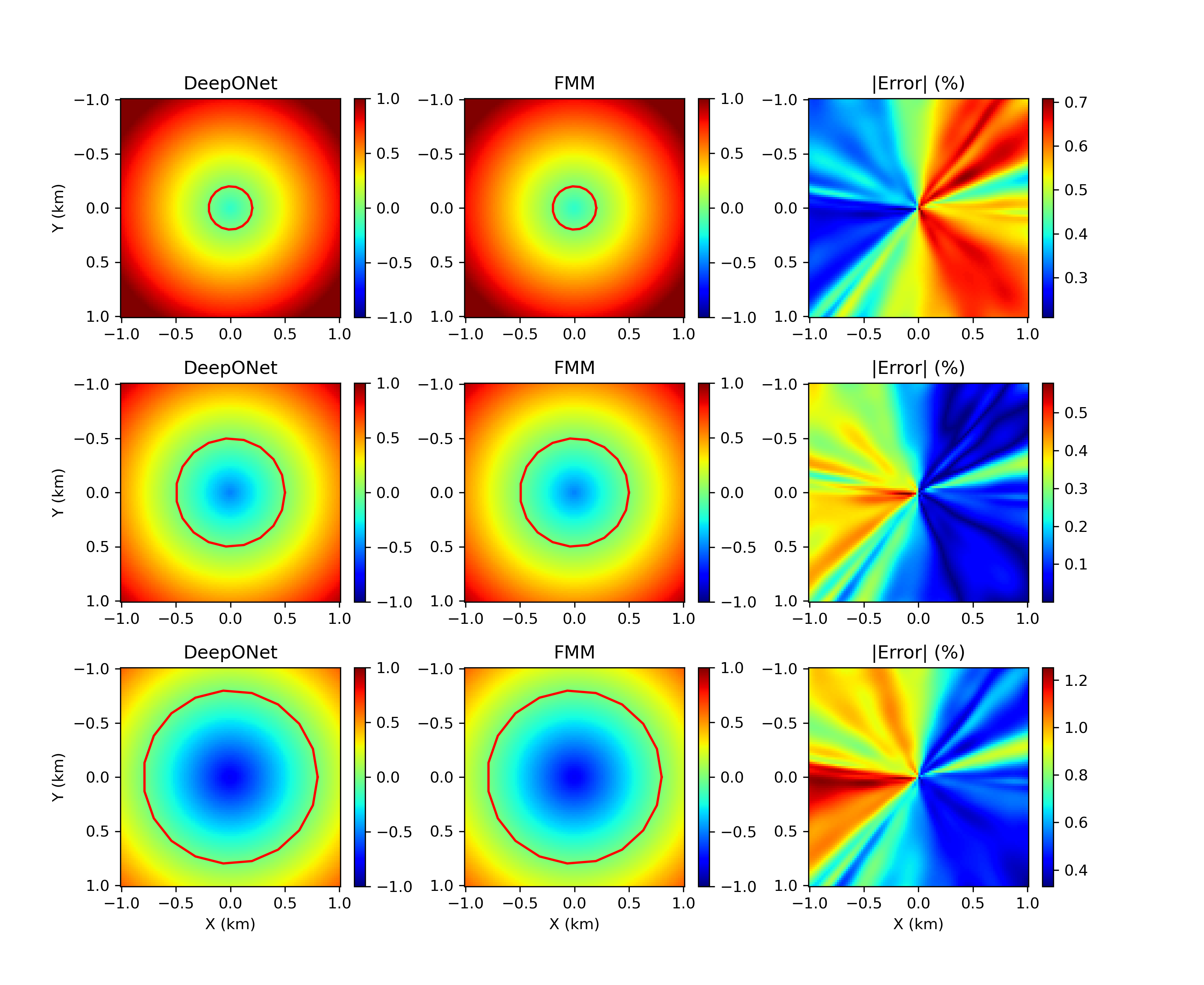}
    \caption{A standard physics-informed DeepONet solution for learning the distance function from a circle with a variable radius. The red circles indicate the coordinates associated with zero distance and are fed as input to the branch network. The standard DeepONet model shows high accuracy in learning the distance function, compared with those computed using the fast marching method.}
    \label{fig:fig2}
\end{figure}

Let us consider the simplest form of the eikonal equation -- the isotropic eikonal equation with a constant velocity model ($v(\bs{x})=1$). This problem can be solved under both data-driven and physics-informed optimizations, see \cite{wang2021learning}, where the authors' objective was learning distance function from a circle with a variable radius. 
The inputs to the branch network are the spatial coordinates $(x, y)$ of a fixed set of nodes located on the circle, which are associated with the zero distance. As we increase or decrease the radius of the circle, $x,y$ values are scaled proportionally. The results are plotted in \cref{fig:fig2}. It can be observed that the standard DeepONet model can accurately capture the solution for various radii, with a maximum error of 1.3\% of the true solution. 

Next, we consider the sensors located on a circle with a fixed radius of $0.2$m but with a moving center to simulate the earthquake source localization problem. As one would expect, for an isotropic medium with a constant velocity model, the emanating wavefront propagates as a circle (in 2D) and only shifts with the location of the hypocenter.
To have a fair comparison, we set up the networks in such a way that the number of trainable parameters are the same for both standard DeepONet and En-DeepONet networks. Accordingly, we use five hyperbolic-tangent layers with 50 neurons width, with an output linear embedding layer of size 50, for both branch and trunk networks. This choice results in 14,800 trainable parameters for the DeepONet architecture. For the case of the En-DeepONet network, we use two layers with 50 neurons width for both branch and trunk networks, followed by a two layer root network with 50 and 20 neurons respectively, resulting in 15,891 trainable parameters. The activation functions are again hyperbolic-tangent function.

\begin{figure} 
    \centering
    \includegraphics[width=1.\textwidth]{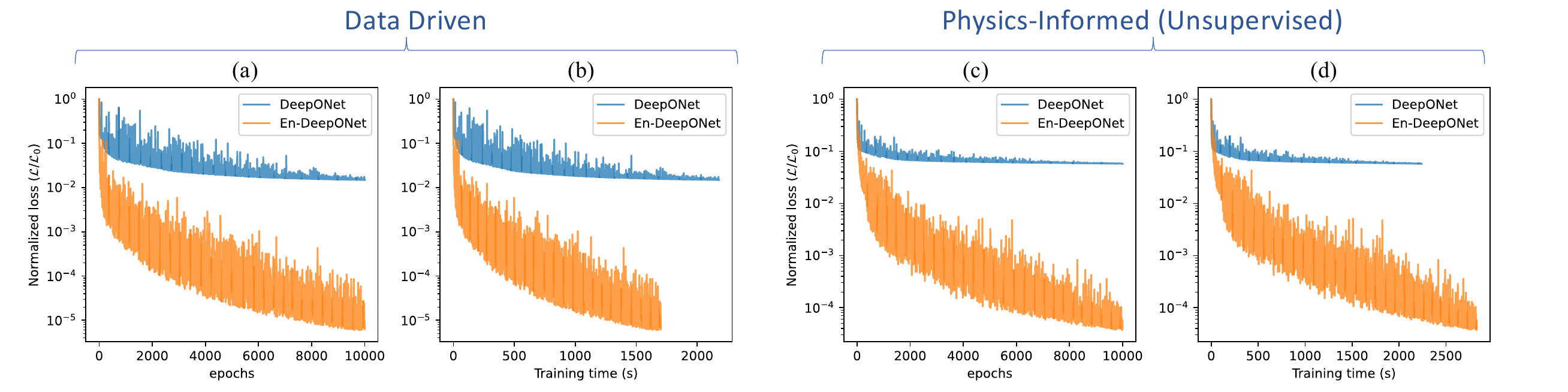}
    \caption{Standard DeepONet vs. En-DeepONet for variable source location with data-driven (a, b) and physics-informed (unsupervised) training (c, d). The vertical axis is the normalized loss function, and the horizontal axes are training epochs (a, c) and training time in seconds (b, d). The training is conducted with a similar number of parameters, the same batch-size and epochs, so that it provides a fair comparison between the two architectures. Note that En-DeepONet loss can be further improved as the loss value is still reducing after 10,000 epochs.}
    \label{fig:fig3}
\end{figure}

\begin{figure} 
    \centering
    \includegraphics[width=0.95\textwidth]{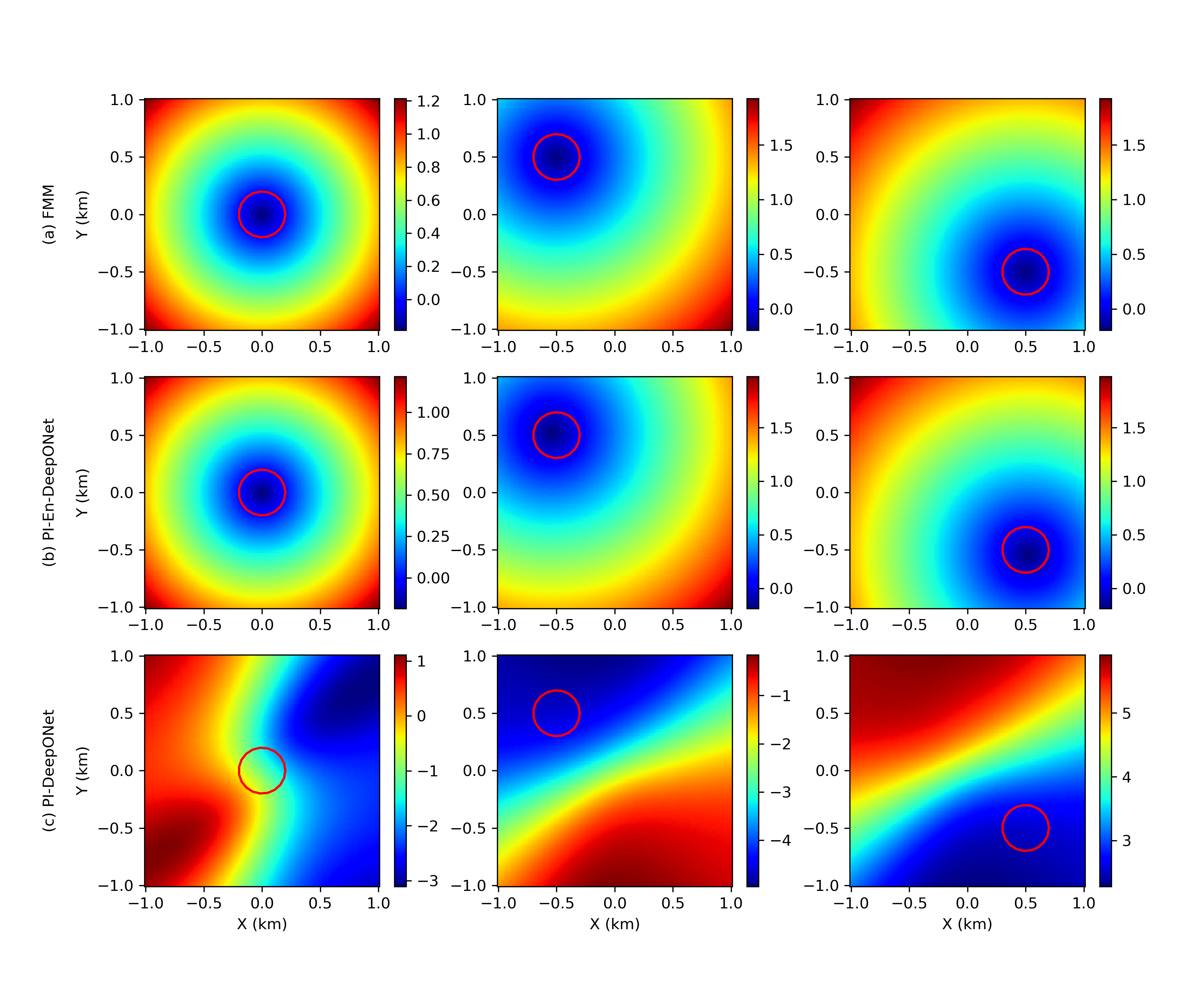}
    \caption{Standard DeepONet vs. En-DeepONet for variable source location with physics-informed (unsupervised) training.  Each column plots the distance field from a randomly located circle, with the horizontal and vertical axes as the $x,y$ coordinates. The plotted solutions correspond to three different locations for the circle, computed using (a) the reference fast marching method (FMM), (b) the physics-informed En-DeepONet, and (c) physics-informed standard DeepONet.
    }
    \label{fig:fig4}
\end{figure}

The results are shown in \cref{fig:fig3}. As can be seen, the standard DeepONet fails to learn the solution under data-driven or physics-informed training, whereas the En-DeepONet accurately captures the expected solution with remarkable accuracy.
Both data-driven and physics-informed (unsupervised) training are performed. The evolution of loss function and training times are shown in \cref{fig:fig3}, where En-DeepONet outperforms standard DeepONet significantly with the same training cost.
The results are shown in \cref{fig:fig4}. As can be seen, the standard DeepONet fails to learn the solution under data-driven or physics-informed training, whereas the En-DeepONet accurately captures the expected solution with remarkable accuracy.

The reason the new architecture outperforms the standard architecture is intuitive. Let us consider the analytical formula for a circular levelset with a fix radius $R$ and centered at the origin, i.e., $\phi(x,y) = x^2 + y^2 - R^2$. Moving the center of the circle to $(x_s, y_s)$ results in a levelset formula that is translated in $(x, y)$, $\phi(x,y) = (x - x_s)^2 + (y - y_s)^2 - R^2$. We can notice that the subtraction operator provides a more natural choice in this case. Now, let us consider a more general case where levelsets form ellipsoids instead of circles, i.e., $\phi(x,y) = a(x - x_s)^2 + b(y - y_s)^2 - R^2$. We can note that the a and b  parameters are now multiplied to the other functions, which is the analogy we use to keep the multiplication operator. Therefore, adding all operators is required to be able to address change in center, anisotropy, and amplitude, and therefore the choice for enriched DeepONet.

This simple problem highlights that the addition or subtraction operators introduced in this paper, in addition to the multiplication operator used in the standard DeepONet construction, is necessary for optimal accuracy when the expected solution shifts with the change in sensor position.

\begin{figure} 
    \centering
    \includegraphics[width=0.75\textwidth]{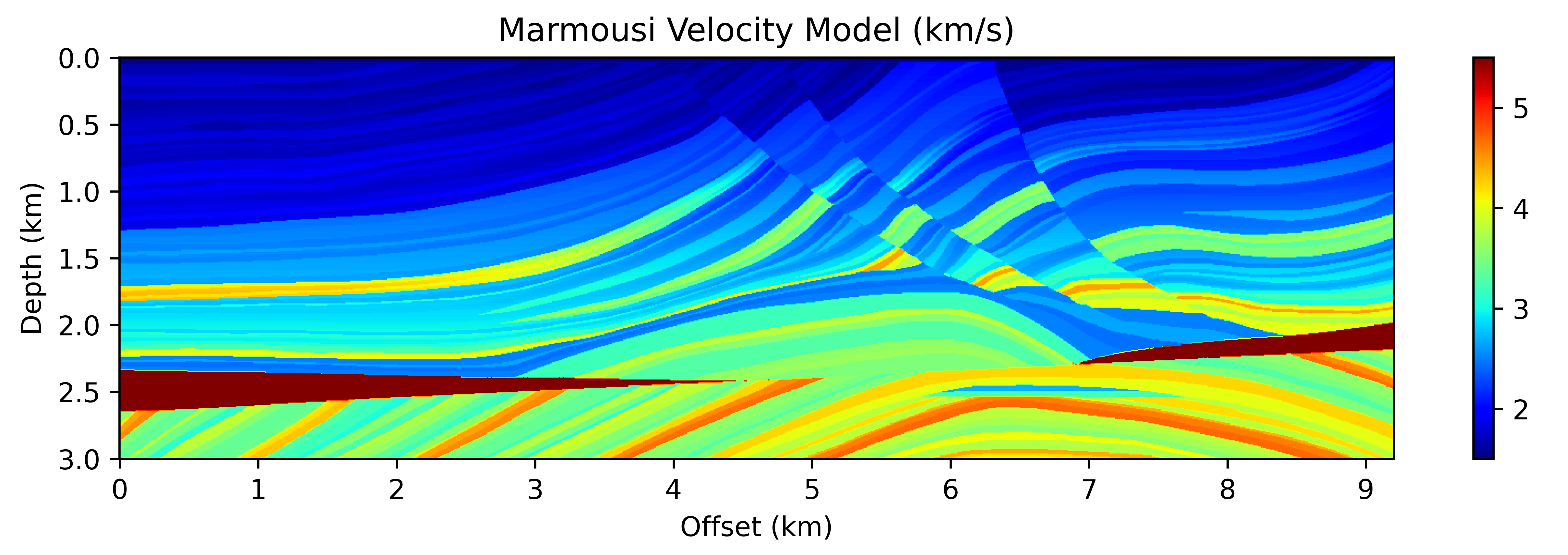}
    \caption{The highly heterogeneous Marmousi velocity model used in this study for assessing the hypocenter localization capabilities of En-DeepONet.}
    \label{fig:fig5}
\end{figure}

\subsection{Source localization for the Marmousi velocity model}

here, we consider a realistic setting for the hypocenter source localization problem by using the highly heterogeneous Marmousi velocity model~\cite{versteeg1994marmousi}, shown in \cref{fig:fig5}. We consider 20 sensors (seismic stations) located at the surface of the velocity model.
The trunk network includes four layers with 20, 50, 100, and 100 neurons. The branch network includes two layers with 50 and 100 neurons. The root network includes 5 hidden layers with 100, 80, 60, 40, 20 neurons. Hyperbolic-tangent is used as the activation function, with a final linear output layer, representing the travel time. We
train En-DeepONet to predict traveltimes in the entire computational domain. The coordinates of the minimum traveltime value then form the hypocenter location.

\begin{figure} 
    \centering
    \includegraphics[width=1\textwidth]{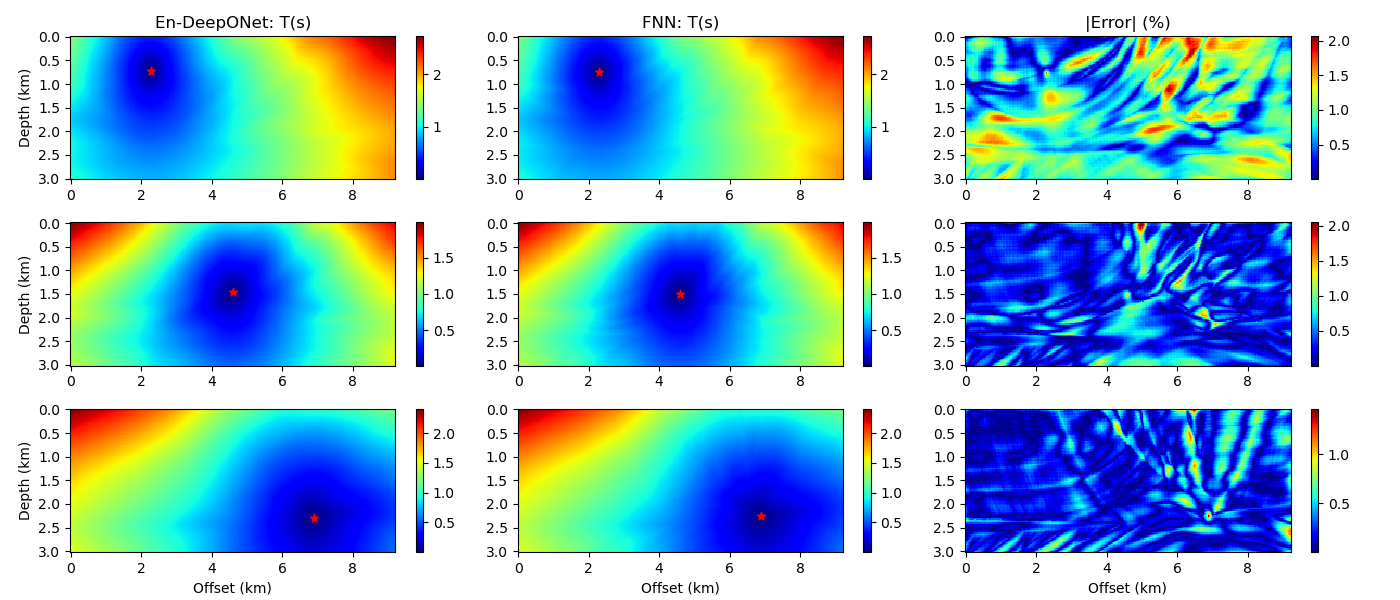}
    \caption{Analyzing the accuracy of traveltime fields predicted by the trained En-DeepONet model for three representative source locations from the test set for the Marmousi model. The En-DeepONet solutions are plotted in the left column, the reference FMM solutions are plotted in the middle column, and the relative errors are plotted in the right column. For each case, the relative error is evaluated as the absolute percentage error normalized by its
 maximum traveltime. The trained En-DeepONet model takes observed traveltimes on the surface as input and produces the traveltime field in the entire domain. The hypocenter is then obtained by the location of the minimum traveltime value, and indicated by the red stars. Despite the heterogeneity of the velocity model, the predicted traveltime field is highly accurate, with maximum error values reaching only 2\%. 
    }
    \label{fig:fig6}
\end{figure}

\begin{figure} 
    \centering
    \includegraphics[width=0.7\textwidth]{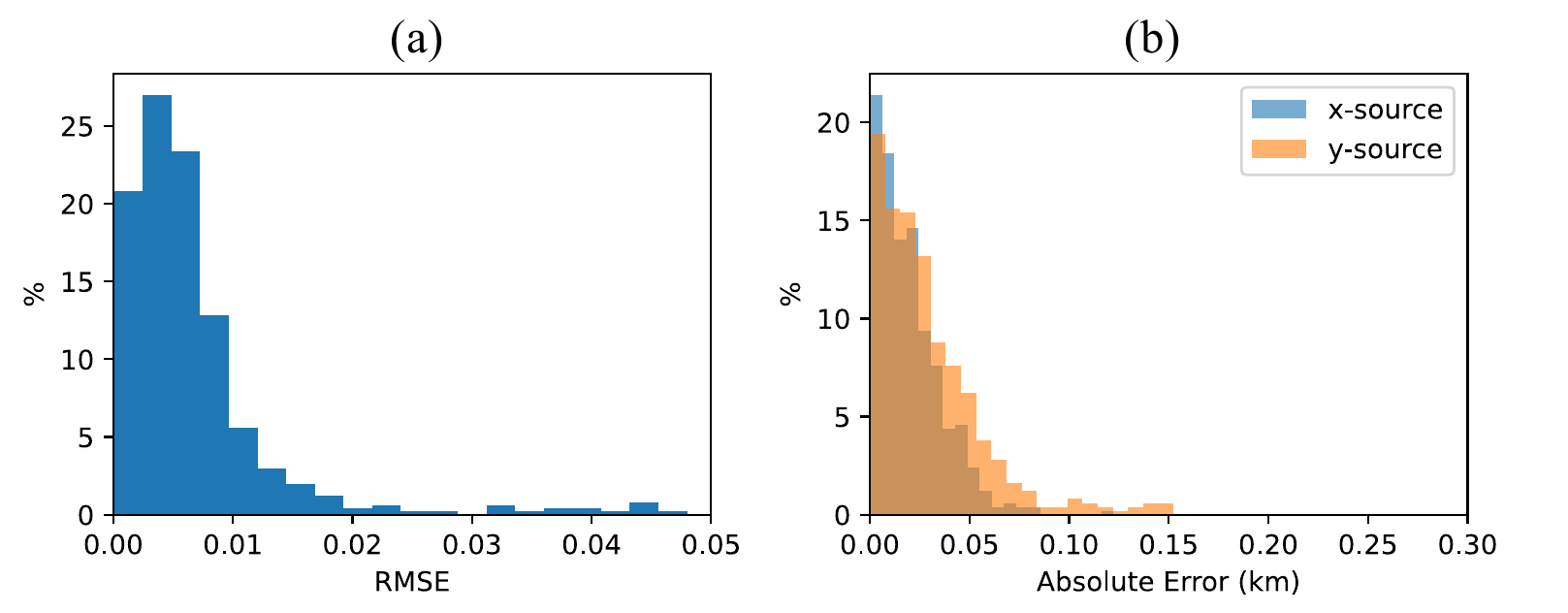}
    \caption{RMSE distribution of predicted traveltimes by the trained En-DeepONet model (a), and a distribution of the absolute error in location coordinates (b), sampled over 1,000 random realizations for the Marmousi model.}
    \label{fig:fig7}
\end{figure}

\begin{figure} 
    \centering
    \includegraphics[width=1.\textwidth]{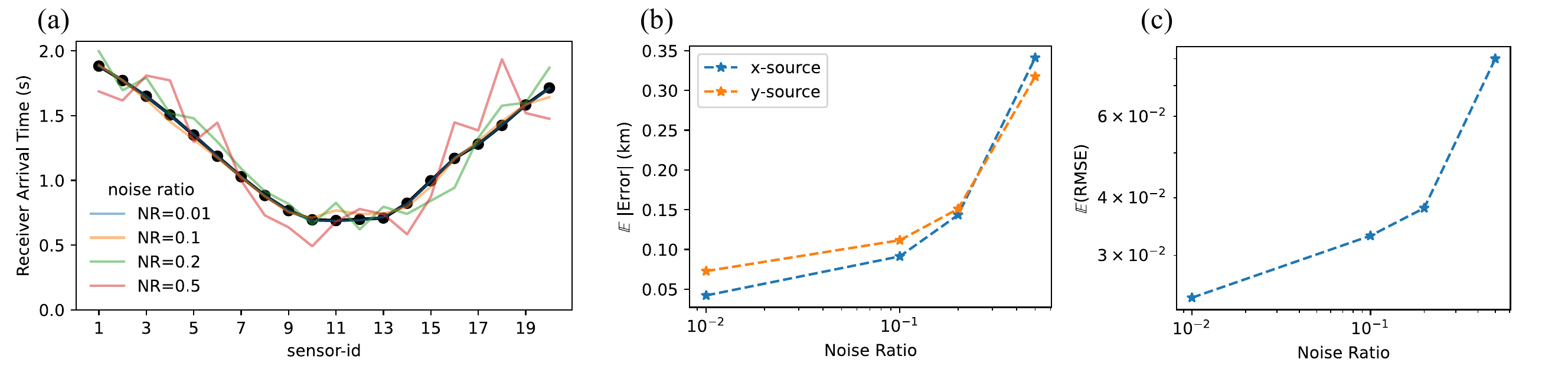}
    \caption{An assessment of the degradation in accuracy of the predicted traveltime fields and the resulting locations due to the addition of noise to the observed traveltime picks. For a representative source location in the Marmousi model, in (a), we plot the observed traveltimes (in black) and the resulting traveltime curves by adding noise with varying ratios. In (b), we plot the evolution of the error in the estimated source locations, averaged over 1,000 random realizations with corrupted traveltime inputs. Finally, in (c), we show the evolution of the RMSE for traveltime predictions, averaged over 1,000 random realizations with the corrupted traveltime inputs. Despite the added noise to the observed traveltimes, the source location is predicted with high accuracy.
    }
    \label{fig:fig8}
\end{figure}

\begin{figure} 
    \centering
    \includegraphics[width=1.\textwidth]{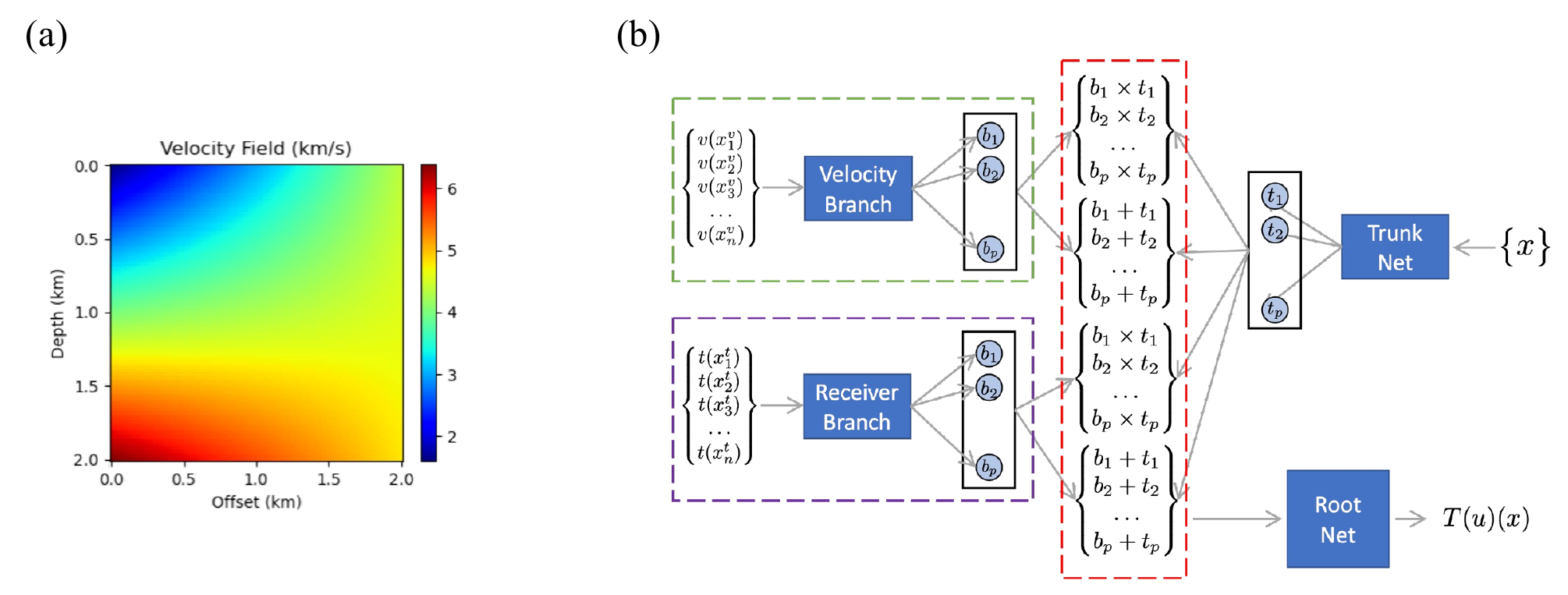}
    \caption{(a) A representative velocity model generated by random sampling of velocity at corners of the square and bilinear interpolation. (b) The En-DeepONet architecture with both velocity and receiver branches, used to generalize over both the source location and the velocity model. We use two branch networks -- one for the velocity model (velocity branch) and the other for the observed traveltimes at the receivers (receiver branch), and a trunk network. These branch and trunk networks are combined with the En-DeepONet rules and then passed as input to the root network. The resulting output is a function of the velocity model sampled at discrete locations, traveltime sampled at discrete receiver points, and spatial coordinates.}
    \label{fig:fig9}
\end{figure}

\begin{figure} 
    \centering
    \includegraphics[width=0.95\textwidth]{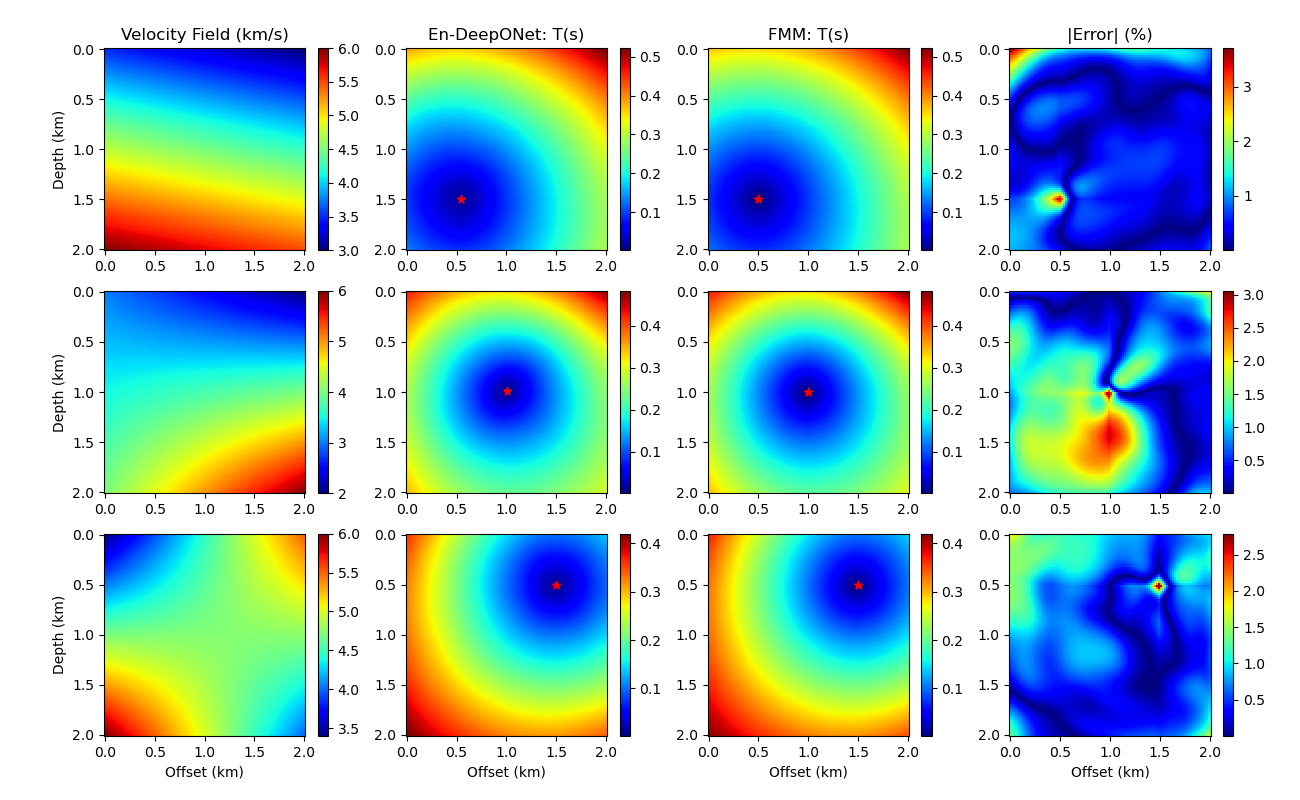}
    \caption{Analyzing the accuracy of traveltime fields predicted by the trained En-DeepONet model for three representative velocity models (first column from the left) and source locations from the test set. Also plotted are the corresponding traveltime fields predicted by the trained En-DeepONet model (second column from the left), the corresponding FMM reference solution (third column from the left), and the resulting absolute relative percentage  error (fourth column from the left). The trained En-DeepONet model takes observed traveltimes on the surface and the velocity model as inputs and produces the traveltime field in the entire domain. The hypocenter is then obtained by the location of the minimum traveltime value, and indicated by the red stars. Despite the challenge of varying velocity models, the predicted traveltime field for a new velocity model and source location both is highly accurate.}
    \label{fig:fig10}
\end{figure}

We generate the training dataset by using the fast marching eikonal solver~\cite{sethian1999fast} for 200 randomly selected source locations within the Marmousi model. Here, the branch net encodes the arrival times observed at seismic stations located on the top surface. Once trained, we test our En-DeepONet model using 1000 randomly selected source locations from within the velocity model. A few representative results are plotted in \cref{fig:fig6}. As can be seen, the solution remains very accurate even for such a heterogeneous velocity model, which shows the potential application of the proposed framework for realistic problems. 

A summary of the results is presented in terms of the root-mean-squared error (RMSE) and absolute error histograms in~\cref{fig:fig7}. We observe that in most cases, the horizontal and vertical location errors are less than 50~m. The vertical location errors are larger on average than the horizontal ones, which is explained by the surface acquisition geometry used in this test. Finally, we study the sensitivity of the location accuracy to noise in the observed arrival times. Through absolute error and RMSE, \cref{fig:fig8} shows the location accuracy remains largely accurate even for an extremely high noise ratio of 50\%.

\subsection{Source and velocity generalization for a smooth velocity model}

As the next representative problem, we now consider a more complex setup. Given an arbitrary velocity model and arrival times at seismic stations located at the top boundary of the velocity model, we aim to build an En-DeepONet model for the hypocenter localization problem.
Here, we consider 20 seismic receivers, equally spaced on the surface. We additionally consider 100 velocity sensors on a uniform grid inside the domain. 
The trunk network includes five layers with 20, 50, 100, 100, and 100 neurons. Both the velocity and the receiver branch networks include three layers with 100 neurons. A linear layer is considered here for the root network. Hyperbolic-tangent is used as the activation function. The final output represents the travel time.

\begin{figure} 
    \centering
    \includegraphics[width=0.7\textwidth]{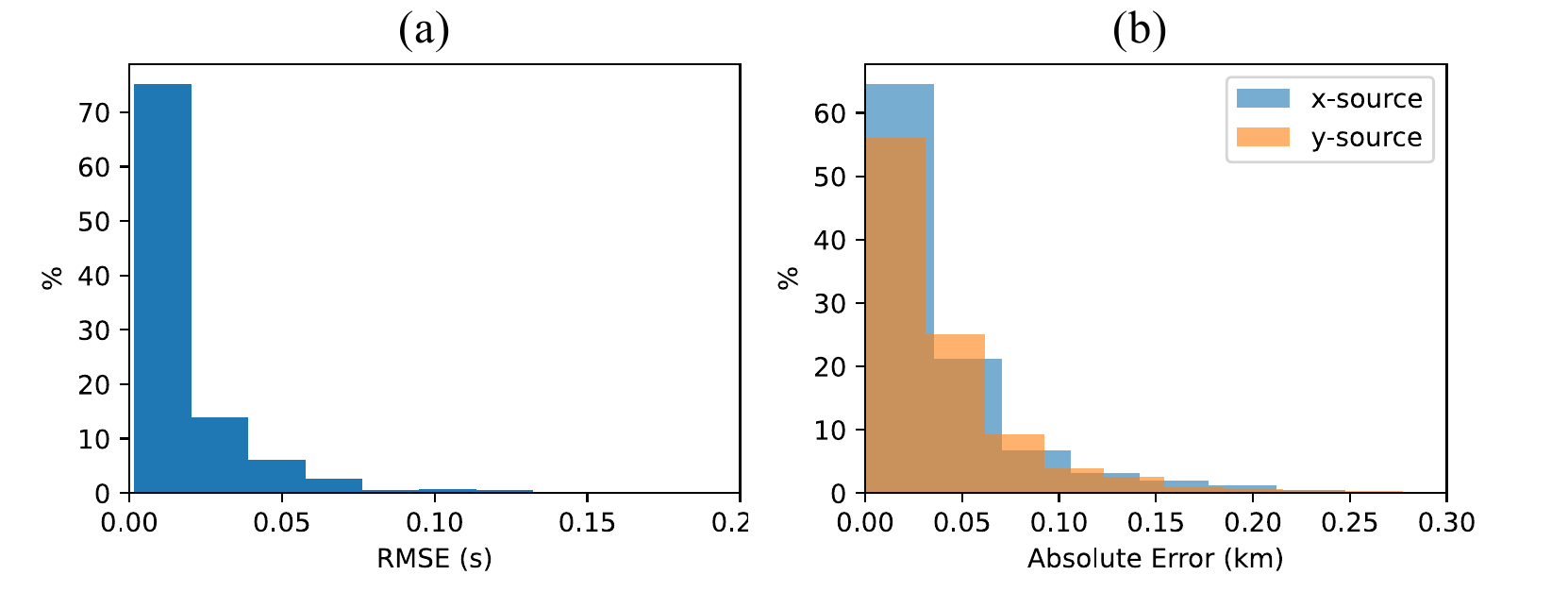}
    \caption{RMSE distribution of predicted traveltimes by the trained En-DeepONet model (a), and distribution of the absolute error in location coordinates (b), sampled over 1,000 random realizations.}
    \label{fig:fig11}
\end{figure}

\begin{figure} 
    \centering
    \includegraphics[width=1.\textwidth]{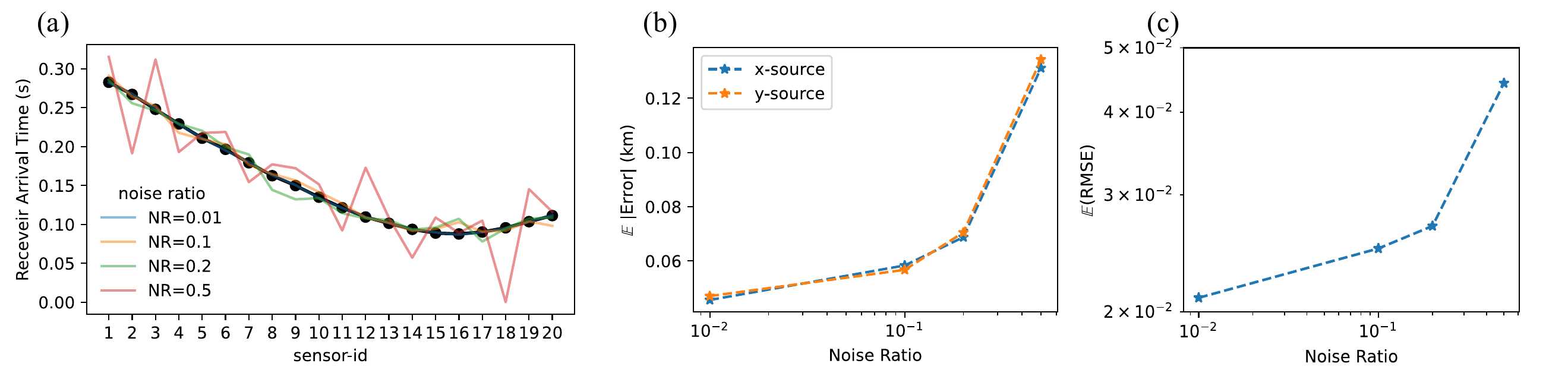}
    \caption{An assessment of the degradation in accuracy of the predicted traveltime fields and the resulting locations due to the addition of noise to the observed traveltime picks. For a representative source location and a velocity model, in (a), we plot the observed traveltimes (in black) and the resulting traveltime curves by adding noise with varying ratios. In (b), we plot the evolution of average error in the estimated source locations over 1,000 random realizations with corrupted traveltime inputs. Finally, we also show in (c) the evolution of the RMSE for traveltime predictions, averaged over 1,000 random realizations with the corrupted traveltime inputs. Despite the complexity of the problem, the solution accuracy does not deteriorate much with the increasing noise levels in the observed traveltimes.}
    \label{fig:fig12}
\end{figure}

To this end, let us consider a square domain, with its corners being assigned random velocities drawn from a representative uniform distribution, i.e., $v \sim \mathcal{U}(0.2, 8)$km/s, as shown in \cref{fig:fig9}-a. The network architecture is depicted in \cref{fig:fig9}-b. Here, we need two branch networks to parameterize the velocity models as well as arrival times. The trunk network remains the same, with spatial coordinates as inputs. 
To generate the training data, we draw a random velocity model, as well as a random source location from $(x_s, y_s) \sim\mathcal{U}(-0.9, 0.9)$m, within the desired domain $\Omega = [-1,1]\times[-1,1]$. Solving the eikonal equation using FMM provides us traveltimes  needed to build the input-output pairs for the training dataset. We generate 200 such samples to train our En-DeepONet model, and then we test it on 1,000 random realizations.

Once trained, given a new velocity model from within the training distribution and wavefront arrival times on the surface, we can instantly query the network to recover the traveltime field and, therefore, the hypocenter location associated with the arrival time and velocity model. The results for a few representative realizations are presented in \cref{fig:fig10}. The trained network is able to predict solutions associated with random source location and velocity models very accurately. It is worth highlighting that the test velocity models are not used during the training. However, as long as they are from within the training distribution, we can expect the trained En-DeepONet model to generate highly accurate solutions.

We consider 1000 randomly generated pairs of velocity models and hypocenter locations to test the localization accuracy. The RMSE of the traveltime field and and absolute location error histograms are shown in \cref{fig:fig11}. We observe extremely small location errors, indicating that our En-DeepONet model is able to generalize well over the considered family of smooth velocity models. Additionally, by corrupting traveltimes observed at receivers using random noise with different ratios (\cref{fig:fig12}-a), we can evaluate the average error in source location and traveltime RMSE as a function of noise ratio, as shown in \cref{fig:fig12}-c-d. We observe that for a large portion of samples, the traveltime predictions and the estimated source location remain highly accurate, underscoring the efficacy of the proposed approach for practical scenarios.

\subsection{Source and velocity generalization for the OpenFWI velocity models}

\begin{figure}
    \centering
    \includegraphics[width=\textwidth]{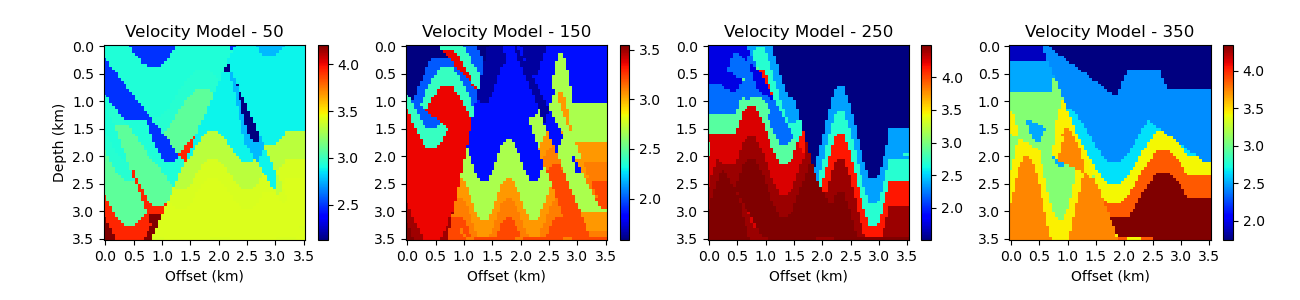}
    \caption{A few representative velocity models from the CurveFault-B family of velocity models in the OpenFWI dataset~\cite{deng2021openfwi}.}
    \label{fig:fig13}
\end{figure}

Finally, as a challenging test case, we consider a family of velocity models from the OpenFWI velocity dataset. OpenFWI represents a comprehensive suite of large-scale, multi-structured datasets, specifically curated for enhancing machine learning applications in seismic Full Waveform Inversion (FWI)~\cite{deng2021openfwi}. OpenFWI provides access to twelve different families of velocity models, in varying degrees of complexity. For our study, we choose the CurveFault-B family of velocity models, representing extremely complex geologic conditions. A few samples from this family of velocity models are shown in~\cref{fig:fig13}.

In this problem, we consider 30 seismic receivers, equally spaced on the surface. We additionally consider $14\times 14$ uniformly distributed velocity sensors inside the domain. 
The trunk network includes nine layers, each with 50 neurons. The velocity branch includes five layers with 100, 80, 50, 50, and 50 neurons, while the receiver branch includes five layers, each with 50 neurons. Finally, the root network includes 8 hidden layers, each with 50 neurons, and a final output linear layer that represents the travel time. The hyperbolic-tangent function is used as the activation function.

\begin{figure}
    \centering
    \includegraphics[width=\textwidth]{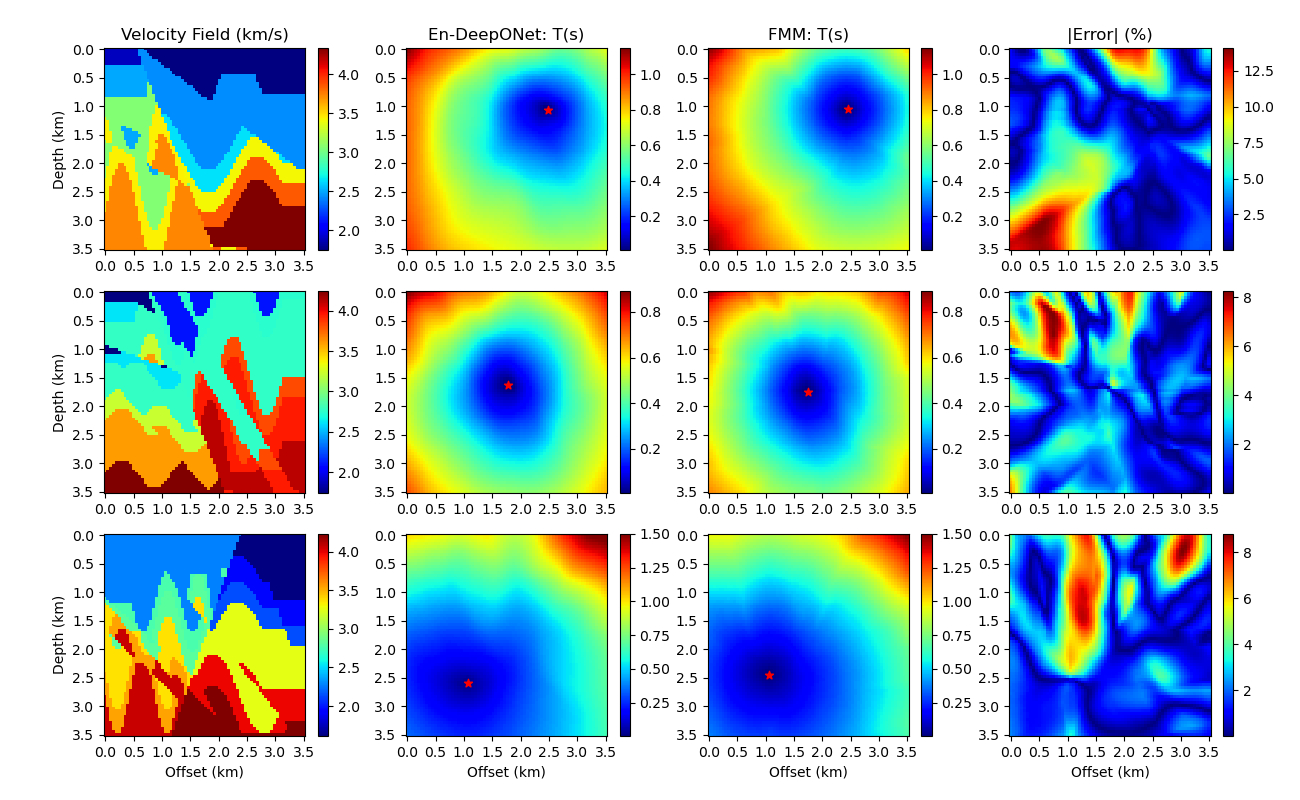}
    \caption{Analyzing the accuracy of traveltime fields predicted by the trained En-DeepONet model for three representative velocity models from the OpenFWI dataset (first column from the left) and source locations from the test set. Also plotted are the corresponding traveltime fields predicted by the trained En-DeepONet model (second column from the left), the corresponding FMM reference solution (third column from the left), and the resulting absolute relative percentage error (fourth column from the left). The trained En-DeepONet model takes observed traveltimes on the surface and the velocity model as inputs and produces the traveltime field in the entire domain. The coordinates of the minimum traveltime value then yield the hypocenter location that are indicated by the red stars. Despite the challenge of dealing with a highly complex set of velocity models, the predicted traveltime fields are generally accurate for most part of the computational domain. }
    \label{fig:fig14}
\end{figure}

\begin{figure} 
    \centering
    \includegraphics[width=0.7\textwidth]{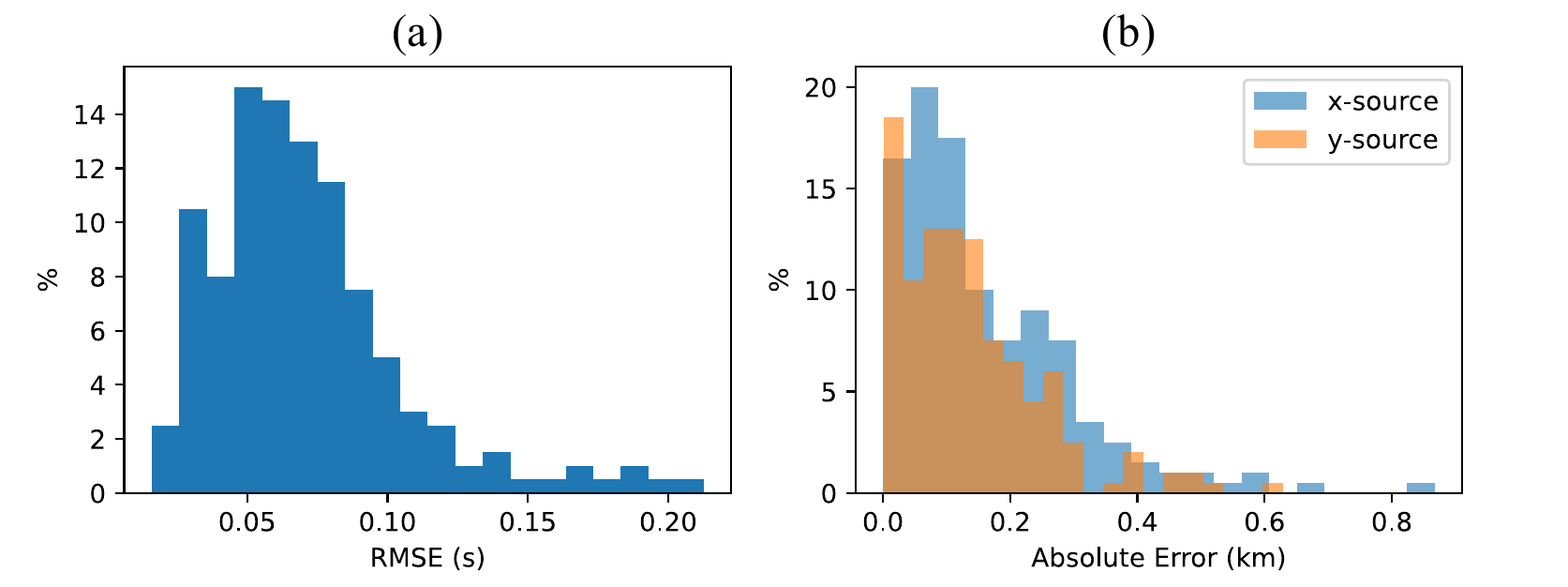}
    \caption{RMSE distribution of predicted traveltimes by the trained En-DeepONet model (a), and distribution of the absolute error in location coordinates (b), sampled over 1,000 random realizations for the CurveFault-B family of velocity models in the OpenFWI dataset.}
    \label{fig:fig15}
\end{figure}

\begin{figure} 
    \centering
    \includegraphics[width=1.\textwidth]{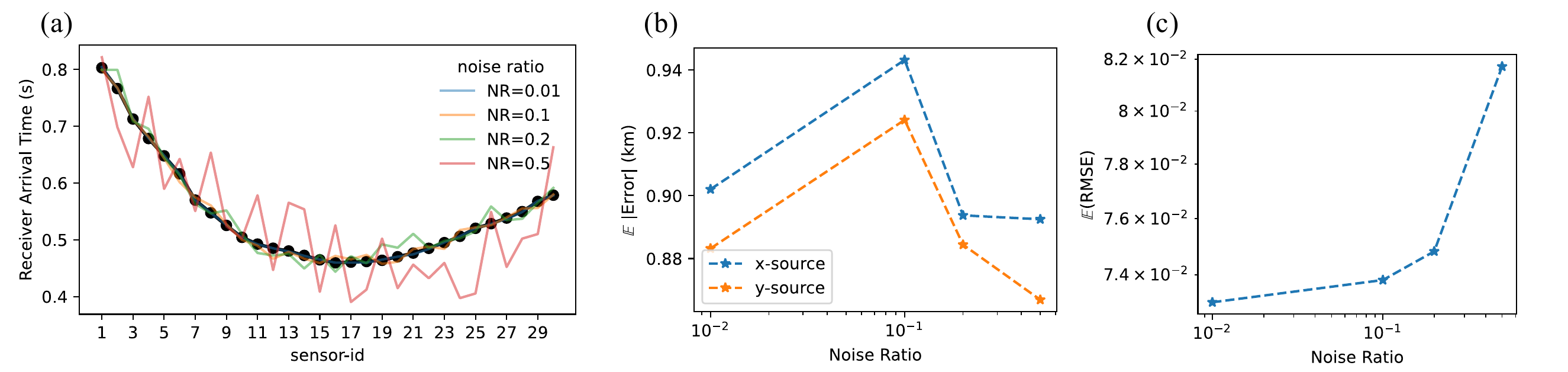}
    \caption{An assessment of the degradation in accuracy of the predicted traveltime fields and the resulting locations due to the addition of noise to the observed traveltime picks. For a representative source location and a velocity model, in (a), we plot the observed traveltimes (in black) and the resulting traveltime curves by adding noise with varying ratios. In (b), we plot the evolution of average error in the estimated source locations over 1,000 random realizations with corrupted traveltime inputs. Finally, we also show in (c) the evolution of the RMSE for traveltime predictions, averaged over 1,000 random realizations with the corrupted traveltime inputs. The solution accuracy is generally lower than earlier but expected due to the complexity of learning an operator corresponding to the complex CurveFault-B family of the OpenFWI dataset.}
    \label{fig:fig16}
\end{figure}

Similar to the previous case, we train the En-DeepONet model (\cref{fig:fig9}-b) to learn the variations in the resulting traveltime field w.r.t. variations in both the observed arrival times and the velocity model. Using FMM, we generate 200 input-output pairs of the training dataset by considering different velocity models from the dataset and source locations. 
Once trained, we evaluate the performance of the En-DeepONet model by feeding new velocity models and the corresponding arrival times observed on the surface from the test set. A few representative results are shown in~\cref{fig:fig14}. 
We notice that despite being from the same family, the velocity models vary significantly compared to each other. Despite the huge differences in velocity models, the trained En-DeepONet model yields a traveltime map with reasonable accuracy for most of the computational domain. While the errors are certainly larger than in the previous test cases, this challenging family of velocity models is deliberately chosen to test the limits of the proposed algorithm. 

We present a summary of these tests through an RMSE distribution of the traveltime field and a histogram of location errors in~\cref{fig:fig15}. While in most cases, the location errors in both the horizontal and vertical directions are less than 200 m, there are some instances in which these errors exceed 500 m, which suggests further improvements may be needed before the proposed approach can be used in critical applications for such a complex family of velocity models.

Finally, in~\cref{fig:fig16}, we show the degradation of accuracy of the predicted traveltime fields and the resulting locations as we pollute the observed arrival times by the addition of random noise with varying ratios. As expected, the situation generally worsens with the addition of noise, resulting in even higher errors in traveltime fields and locations.

\section{Summary and conclusions}

In this study, we introduced En-DeepONet, an enhanced variant of deep operator networks, for near real-time earthquake hypocenter localization. By incorporating information from seismic arrival times and velocity models, En-DeepONet effectively estimated traveltime fields associated with earthquake sources. The addition of the root network, which included multiplication, addition, and subtraction operators, improved the accuracy of field relocation in scenarios with moving fields. 
Through extensive experiments, including various velocity models and arrival times, En-DeepONet demonstrated remarkable accuracy in earthquake localization, even in complex and heterogeneous environments. The framework exhibited excellent generalization capabilities and robustness against noisy arrival times. The method offers a computationally practicable method to quantify uncertainty in hypocenter locations due to errors in traveltime picks and velocity model variations.
Moreover, our findings highlight the potential of En-DeepONet to enhance seismic monitoring systems and contribute to the development of effective early warning systems for mitigating seismic hazards.

In this article, we focused on a two-dimensional velocity model to illustrate the efficacy of En-DeepONet compared to the standard DeepONet architecture. Extension of the idea to three-dimensions is essential for solving problems of practical interests in hypocenter localization. Given that the standard DeepONet architecture has proven to outperform other operator networks, like Fourier Neural Operators, in three dimensions, we believe that our proposed En-DeepONet architecture further improves the representation capabilities of DeepONets without any additional computational cost. Moreover, a refinement of the proposed approach may be needed in the case when traveltime picks from certain sensors are unavailable during evaluation of the trained En-DeepONet.

\section*{Acknowledgments}

We acknowledge the support provided by the College of Petroleum Engineering and Geosciences (CPG) at King Fahd University of Petroleum \& Minerals (KFUPM) through Competitive Research Grant No. CPG21103.

\section*{Data availability}
The data and codes used in this study will be shared at \href{https://github.com/ehsanhaghighat/En-DeepONet}{https://github.com/ehsanhaghighat/En-DeepONet}.

\appendix
\section{Application to the Laplace equation}

To show that the En-DeepONet architecture is not limited to the Eikonal equation, we consider the Laplace equation in two dimensions subjected to a moving pulse source term, representing injection/production wells in porous media or a moving heat source term. The governing PDE is expressed as 
\begin{equation}
    \frac{\partial^2 u(\bs{x})}{\partial x^2} + \frac{\partial^2 u(\bs{x})}{\partial y^2} = -\exp\left(-\frac{\|\bs{x}-\bs{x}_s\|^2}{\delta^2}\right), \quad \bs{x} \in \Omega,
\end{equation} 
subjected to the boundary conditions
\begin{equation}
    u(\bs{x}) = 0, \quad \bs{x} \in \partial \Omega,
\end{equation}
where, $\bs{x}_s$ is the center of the source location, and $\delta$ is the source width. We consider a square domain $\Omega = [-1,1]\times[-1,1]$, and we consider a fixed source width $\delta^2 = 0.1$. The source center $\bs{x}_s$ is chosen randomly from $\bs{x}_s \sim \mathcal{U}(-0.9, 0.9)$m to build the neural operator. We only consider unsupervised (physics-informed) in this case, meaning that the training is only based on the PDE and its boundary condition, without using any other data.

The input to the branch net is the location of the center of the pulse, i.e., $\bs{x}_s$, therefore sensor size is 2 in two dimensions. As per the network details, to provide a fair comparison between standard DeepONet and En-DeepONet architectures, we keep total trainable parameters similar for both cases. Accordingly, for the standard DeepONet network, we use five hyperbolic-tangent layers with 50 neurons width, with an output linear embedding layer of size 50, for both branch and trunk networks. This choice results in 12,900 trainable parameters. For the case of the En-DeepONet network, we use two layers with 50 neurons width for both branch and trunk networks, followed by a two layer root network with 50 and 20 neurons respectively, resulting in 13,900 trainable parameters. The activation functions are again hyperbolic-tangent function. The input/output sampling points and optimization hyper parameters are kept identical for both cases. 

\begin{figure} 
    \centering
    \includegraphics[width=0.95\textwidth]{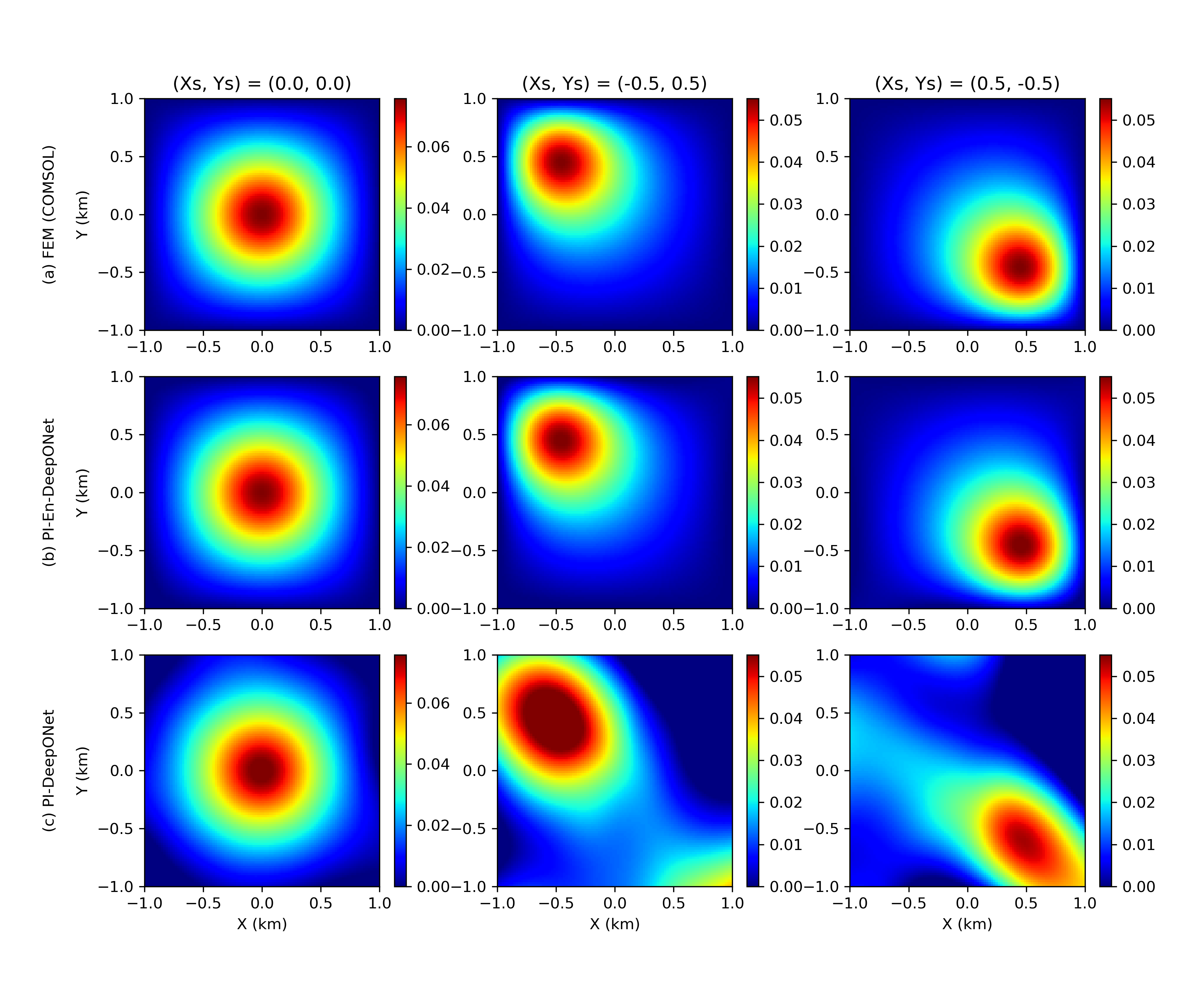}
    \caption{Standard DeepONet vs. En-DeepONet for variable source location for the Laplace equation, with physics-informed (unsupervised) training.  Each column plots the temperature field force a randomly selected source location, with the horizontal and vertical axes as the $x,y$ coordinates. The plotted solutions correspond to three different locations for the center of the source pulse, computed using (a) the reference solution from Finite Element Method (FEM) using COMSOL software, (b) the physics-informed En-DeepONet, and (c) physics-informed standard DeepONet.}
    \label{fig:fig17}
\end{figure}

\begin{figure} 
    \centering
    \includegraphics[width=0.6\textwidth]{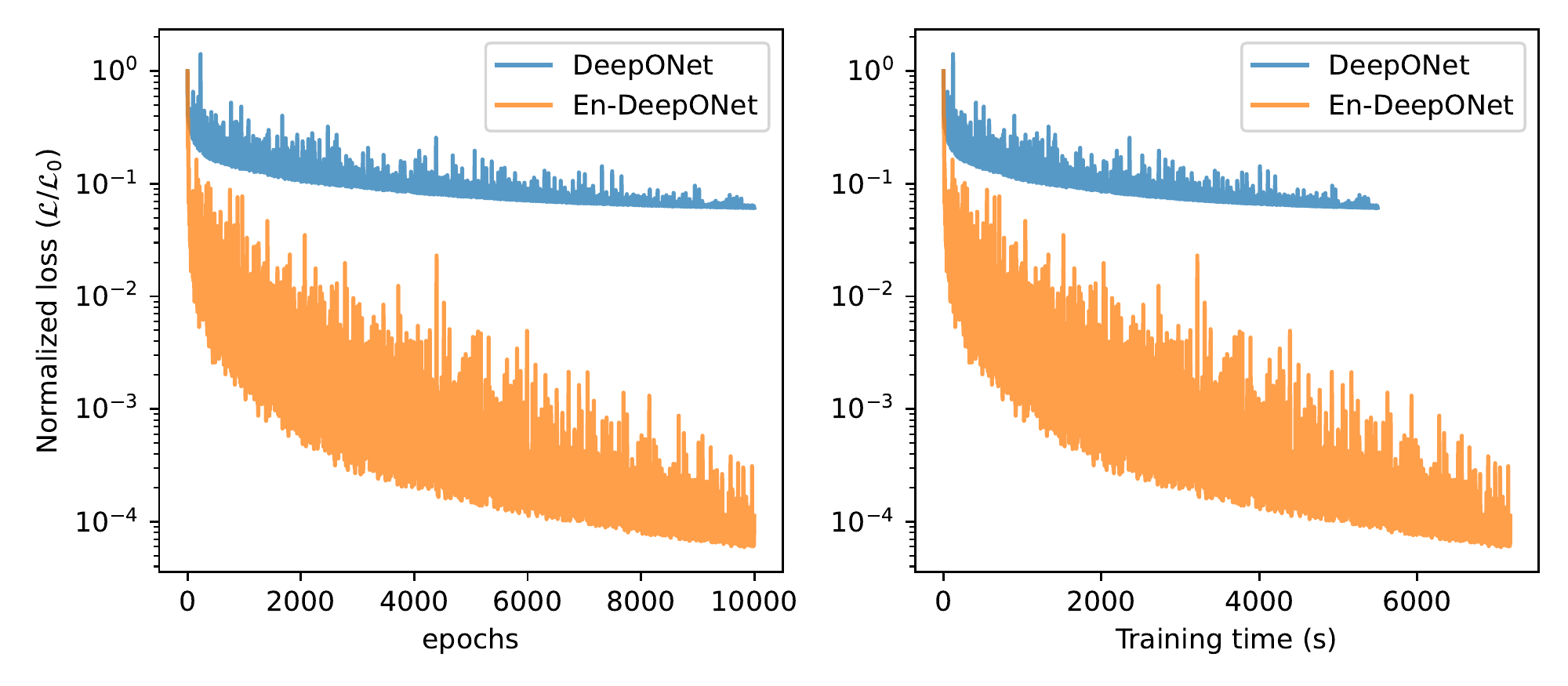}
    \caption{Loss function for the standard DeepONet vs. En-DeepONet for moving source location for the Laplace equation, with physics-informed (unsupervised) training. The vertical axis is the normalized loss function, and the horizontal axes are training epochs (Left) and training time in seconds (Right). The training is conducted with a similar number of parameters, the same batch-size and epochs, so that it provides a fair comparison between the two architectures. Note that En-DeepONet loss can be further improved as the loss value is still reducing after 10,000 epochs.}
    \label{fig:fig18}
\end{figure}

The results are shown in \cref{fig:fig17}. As can be seen, the standard DeepONet fails to learn the solution operator for this moving source problem, whereas the En-DeepONet accurately captures the expected solution, which is an identical trend compared to those previously reported for the Eikonal equation (see \cref{fig:fig4}). The loss history \cref{fig:fig18} also shows a similar trend to those previously reported in \cref{fig:fig3}.

\bibliographystyle{elsarticle-num} 
\bibliography{references}

\end{document}